\documentclass[runningheads]{llncs}
\usepackage[T1]{fontenc}
\usepackage{graphicx}
\usepackage{amssymb}
\usepackage{mathdots}
\usepackage{stmaryrd}
\usepackage{amsmath}
\usepackage{booktabs}
\usepackage{longtable}
\usepackage{xcolor,colortbl}
\usepackage{multirow, colortbl, xcolor, booktabs}
\usepackage{soul}
\usepackage{bm}
\usepackage{marvosym}

\definecolor{airforceblue}{rgb}{0.36, 0.54, 0.66}

\begin{document}
\title{S$^3$R: Self-supervised Spectral Regression for Hyperspectral Histopathology Image Classification}
\titlerunning{S$^3$R: Self-supervised Spectral Regression}

\author{Xingran Xie \and
Yan Wang\textsuperscript{(\Letter)} \and
Qingli Li}
\authorrunning{X. X et al.}
\institute{{Shanghai Key Laboratory of Multidimensional Information Processing, \\East China Normal University, Shanghai 200241, China}\\
\email{51215904112@stu.ecnu.edu.cn,}
\email{ywang@cee.ecnu.edu.cn, qlli@cs.ecnu.edu.cn}}
\maketitle              
\begin{abstract}
Benefited from the rich and detailed spectral information in hyperspectral images (HSI), HSI offers great potential for a wide variety of medical applications such as computational pathology. But, the lack of adequate annotated data and the high spatiospectral dimensions of HSIs usually make classification networks prone to overfit. Thus, learning a general representation which can be transferred to the downstream tasks is imperative. To our knowledge, no appropriate self-supervised pre-training method has been designed for histopathology HSIs. In this paper, we introduce an efficient and effective Self-supervised Spectral Regression (S$^3$R) method, which exploits the low rank characteristic in the spectral domain of HSI. More concretely, we propose to learn a set of linear coefficients that can be used to represent one band by the remaining bands via masking out these bands. Then, the band is restored by using the learned coefficients to reweight the remaining bands. Two pre-text tasks are designed: (1) S$^3$R-CR, which regresses the linear coefficients, so that the pre-trained model understands the inherent structures of HSIs and the pathological characteristics of different morphologies; (2) S$^3$R-BR, which regresses the missing band, making the model to learn the holistic semantics of HSIs. Compared to prior arts \emph{i.e.}, contrastive learning methods, which focuses on natural images, S$^3$R converges at least 3 times faster, and achieves significant improvements up to 14\% in accuracy when transferring to HSI classification tasks. 

\end{abstract}
%
%
%
\section{Introduction}
Histopathology plays an important role in the diagnostic and therapeutic aspects of modern medicine \cite{Zhang2019Multidimensional,Salvi2021pathology}. As artificial intelligence is evolving rapidly, deep learning based image processing techniques have been extensively reported for histopathological diagnosis \cite{Salvi2021pathology}. Nowadays, the microscopy hyperspectral imaging system has become an emerging research hotspot in the field of medical image analysis \cite{Wang2018HSI}, benefited from the rich spatiospectral information provided by hyperspectral images (HSI). Thus, it provides a new perspective for computational pathology and precision medicine. 

Supervised learning for histopathology image analysis requires a large amount of annotations \cite{L2022MedicalSSL}. Since both expertise and time are needed, the labeled data with high quality annotations are usually expensive to acquire. This situation is more conspicuous for annotating HSIs \cite{Zheng2022Spectral-Spatial}, whose appearances are different compared with RGB images, so pathologists may take longer time in recognizing cancer tissues on HSIs. Typically, an HSI is presented as a hypercube, 
such high spatiospectral dimensions make it difficult to perform accurate HSI classification, especially when the annotated data is limited, which may lead to overfitting problems. Thus, an appropriate HSI pre-training method is imperative.

In recent years, self-supervised pre-training has been successful in both natural language processing \cite{Devlin2019BERT} and computer vision. Previous research of self-supervised learning mainly focuses on contrastive learning, \emph{e.g.}, MOCO \cite{He2020MOCO}, SimCLR \cite{Chen2020SimCLR}, BYOL \cite{Jean2020BYOL} and SimSiam \cite{Chen2021SimSiam}, training encoders to compare positive and negative sample pairs. Among these methods, BYOL \cite{Jean2020BYOL} and SimSiam \cite{Chen2021SimSiam} achieve higher performance than previous contrastive methods without using negative sample pairs. But, the training setup \emph{e.g.}, batch size in original papers is not always affordable for research institutions. 
More recently, masked image modeling (MIM) methods represented by MAE \cite{He2021MAE} have been proved to learn rich visual representations and significantly improve the performance on downstream tasks \cite{Chen2021Maskfeat}. After randomly adding masks to input images, a pixel-level regression target is set as a pretext task. Almost all the MIM methods are designed based on transformers which receive and process tokenized inputs.

Different from natural images, HSIs analyze how light transmittance varies on different forms of tissues or cells, which infect the brightness of various regions \cite{Fu2018HSIrecovery}. This generates dozens even hundreds of narrow and contiguous spectral bands in the spectral dimension. Understanding the inherent spectral structure in self-supervised pre-training is non-trivial for the networks to conduct downstream tasks. To the best of our knowledge, there is not any self-supervised method designing the architecture tailored for microscopy HSI classification yet. In this work, we present \textbf{S}elf-\textbf{S}upervised \textbf{S}pectral \textbf{R}egression (S$^3$R), an efficient and effective pre-training objective that takes advantage of low rankness in the spectral domain. More specifically, we assume that one band can be represented as a linear combination of the remaining bands, and propose to learn the linear coefficients by a Convolutional Neural Network (CNN) based backbone via masking out the remaining bands. We propose two alternative pretext tasks, which have different focuses. (1) \textbf{C}oefficients \textbf{R}egression (S$^3$R-CR), which makes the network to directly regress the ``groundtruth'' linear coefficients learned beforehand. In this way, the pre-trained model acquires an adequate understanding of complex spectral structures and pathological characteristics of different morphologies. (2) \textbf{B}and \textbf{R}egression (S$^3$R-BR), regressing the pixel intensity of the selected band by re-weighting the remaining bands. Since detailed information (edges, textures, etc.) is already stored in different bands of the HSI, our spectral regression approach will allow the model to focus more on the intrinsic association between spectral bands and the overall semantics of the masked image. Experiments show that our S$^3$R-CR and S$^3$R-BR perform better than contrastive learning and masked image modeling methods, and significantly reduces the cost of self-supervised task due to the faster convergence\vspace{-0.3em}.

\begin{figure}[!t]
\begin{center}
\includegraphics[width=0.9\textwidth]{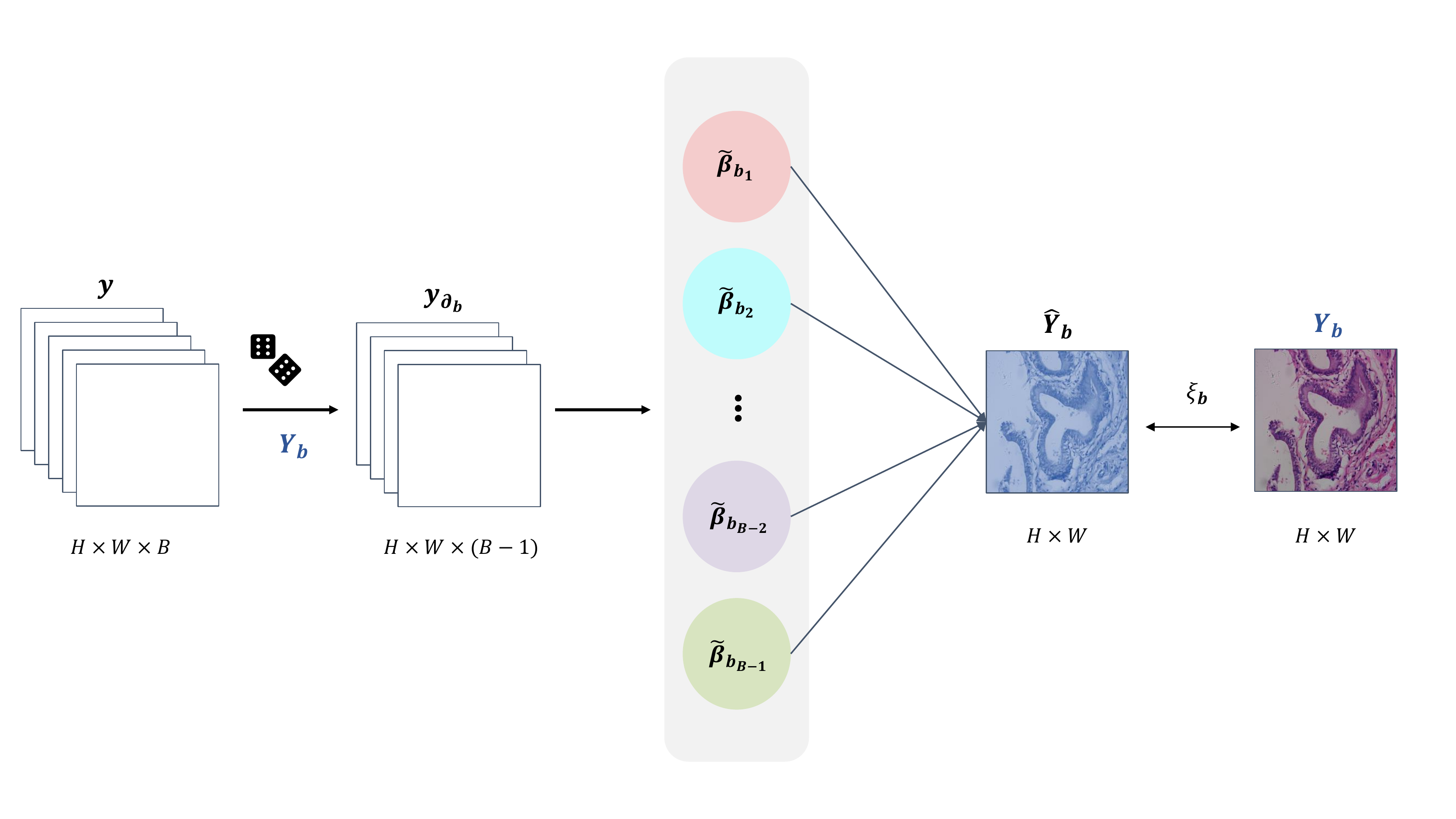}
\end{center}
\caption{Illustration of spectral regression in HSI. The whole process to estimate the coefficient  $\tilde{\bm{\beta}}_b$ can be optimized by back-propagation. $\hat{\mathbf{Y}}_b$ denotes the regression result. The dice icon refers to randomly sampling a band $\mathbf{Y}_b$ as the regression target.} \label{fig:spec_regression}
\end{figure}

\section{Method}

\subsection{Spectral Regression based on Low-rank Prior}{}\label{sec:prior}

Since the spectral correlation in HSIs is extremely high, HSIs can be well represented in a low-dimensional subspace \cite{Zhuang2021fasthymix}. Let $\mathcal{Y}\in\mathbb{R}^{H\times W\times B}$ denote a HSI presented in 3D tensor with $H\times W$ pixels and $B$ spectral bands. We assume that $\mathbf{Y}_b\in\mathbb{R}^{H\times W}$, denoting the $b$-th spectral band in $\mathcal{Y}$, can be represented by a linear combination of the remaining $B-1$ bands with a set of coefficients. This can be formulated as:
\begin{equation}
\label{eq:linear}
    \mathbf{Y}_b = \bm{\xi}_b + \sum_{i=1}^{B}\mathbf{1}[i\neq b]\cdot\mathbf{Y}_i\cdot\beta_{b_i},
\end{equation}
where $\beta_{b_i}$ denotes the linear coefficient corresponding to the $i$th band, and $\mathbf{1}[\cdot]$ is the indicator function. Let $\bm{\beta}_b = (\beta_{b_1},\ldots,\beta_{b_{b-1}},\beta_{b_{b+1}},\ldots,\beta_{b_{B}})^\intercal  \in \mathbb{R}^{(B-1)\times 1}$. Due to the existence of noise from hyperspectral imaging system, the linear regression process will generate a small error, $\bm{\xi}_b \in \mathbb{R}^{H\times W}$, which could not be eliminated completely. Eq.~\ref{eq:linear} provides a linear regression model, which regresses a spectral band by the remaining bands.

To find $\beta_{b_i}$, Eq.~\ref{eq:linear} can be estimated by minimizing the following loss function: 
\begin{equation}
    \mathcal{L} = \left \| \mathbf{Y}_b-\sum_{i=1}^{B}\mathbf{1}[i\neq b]\cdot\mathbf{Y}_i\cdot\beta_{b_i}\right \| _F^2.
\end{equation}
One can simply obtain an estimation of the coefficient $\tilde{\beta}_{b_i}$ through back-propagation by deriving the partial derivatives of $\beta_{b_i}$:

\begin{equation}
\tilde{\beta}_{b_i} = \beta_{b_i} - \alpha\frac{\partial \mathcal{L}}{\partial \beta_{b_i}},
\label{eq:gt_beta}
\end{equation}
or applying a closed-form solution. Let $\tilde{\bm{\beta}}_b = (\tilde{\beta}_{b_1},\ldots,\tilde{\beta}_{b_{b-1}},\tilde{\beta}_{b_{b+1}},\ldots,\tilde{\beta}_{b_{B}})^\intercal  \in \mathbb{R}^{(B-1)\times 1}$. The overall process is shown in Fig.~\ref{fig:spec_regression}.

\begin{figure}[!t]
\includegraphics[width=\textwidth]{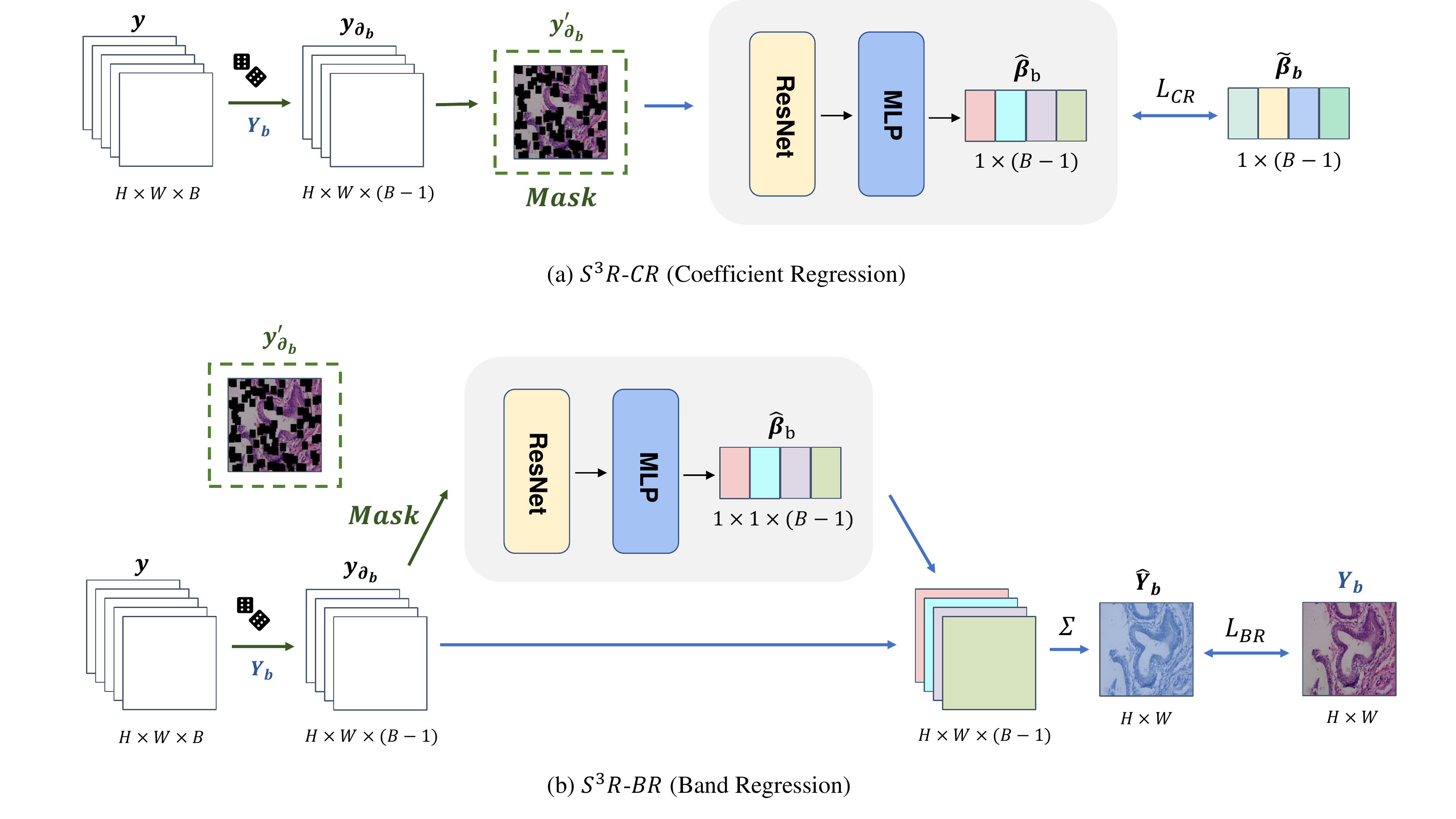}
\caption{\textbf{Our S$^3$R architecture}. During pre-training, we randomly select the $b$th band $\mathbf{Y}_b$ from the input HSI $\mathcal{Y}$. Then the remaining bands $\mathcal{Y}_{\partial_b}$ are masked out by random patches. The encoder is applied to learn a set of coefficients $\hat{\beta}_{b_i}$ from masked bands $\mathcal{Y}'_{\partial _b}$. The learned coefficients are then directly fit $\tilde{\bm{\beta}}_b$ (a) or reweight $\mathcal{Y}_{\partial _b}$ to regress the initially selected band $\mathbf{Y}_b$ (b).} \label{fig:architecture}
\end{figure}

\subsection{Model Based Spectral Regression\label{sec:2.2}}
Previous section indicates that tensor low-rank prior is an inherent property of HSI which does not rely on any supervisory information. Inspired by the low-rank prior, we propose a Self-supervised Spectral Regression (S$^3$R) architecture for hyperspectral histopathology image classification.

As shown in Fig.~\ref{fig:architecture}, we first randomly extract a band $\mathbf{Y}_b$ from an HSI $\mathcal{Y}$. Let $\mathcal{Y}_{\partial _b}\in\mathbb{R}^{H\times W\times (B-1)}$ denote all bands except the $b$th band. Next, we randomly mask out $\mathcal{Y}_{\partial _b}$ to obtain the masked images $\mathcal{Y}'_{\partial _b}$. Then, a CNN based backbone is used to encode and learn a set of coefficients $\hat{\beta}_{b_i}$ given $\mathcal{Y}'_{\partial _b}$. Last, the learned coefficients $\hat{\bm{\beta}}_b$ are applied to a pretext task. 

In our architecture, we consider two pre-training objectives: (1) coefficient regression (S$^3$R-CR) and (2) spectral band regression (S$^3$R-BR), whose details will be given below.

\subsubsection{Image Masking} 
We randomly extract a band $\mathbf{Y}_b\in\mathbb{R}^{H\times W}$ from input HSI $\mathcal{Y}\in\mathbb{R}^{H\times W\times B}$, and mask the remaining bands $\mathcal{Y}_{\partial _b}$ with an approximately 65\%  masking ratio. Like other MIM methods \cite{Li2021BEiT}, manually adding strong noise interference can significantly make the pretext task more challenging. Note that contents in different bands are masked out independently, which is not the same as MIM.

\subsubsection{Coefficients Regression} 
The obtained $\tilde{\bm{\beta}}_{b}$ from Eq.~\ref{eq:gt_beta} can be treated as the groundtruth of the coefficients. Forcing the network to learn the coefficients via randomly masking out a portion of $\mathcal{Y}_{\partial_b}$ will make the network to understand the inherent structure of HSI. Residual architecture is proven to work well for HSI classification \cite{Mou2018Conv-Deconv}. Thus, a vanilla ResNet is adopted to encode the masked tensor $\mathcal{Y}'_{\partial_b}\in\mathbb{R}^{H\times W\times (B-1)}$, followed by a MLP head which maps features to the predicted coefficient $\hat{\bm{\beta}}_{b}$, where $\hat{\bm{\beta}}_b = (\hat{\beta}_{b_1},\ldots,\hat{\beta}_{b_{b-1}},\hat{\beta}_{b_{b+1}},\ldots,\hat{\beta}_{b_{B}})^\intercal  \in \mathbb{R}^{(B-1)\times 1}$. Noted that the MLP head will not be involved in the downstream task. The loss function for optimizing coefficient regression is computed by:

\begin{equation}
\label{eq:cr}
    \mathcal{L}_{CR} = \sum_{i=1}^{B-1} \left \| \hat{\beta}_{b_i} - \tilde{\beta}_{b_i} \right \| _1.
\end{equation}

\subsubsection{Spectral Band Regression} 
As mentioned in Sec.~\ref{sec:prior}, we assume that one band can be represented as a linear combination of the remaining bands. The selected band $\mathbf{Y}_b$ can be represented by the generated linear coefficients $\hat{\bm{\beta}}_b$ and $\mathcal{Y}_{\partial_b}$. 
Then, we aim to minimize the following loss function:

\begin{equation}
\label{eq:br}
    \mathcal{L}_{BR} = \left \| \mathbf{Y}_b - \sum_{i=1}^{B}\mathbf{1}[i\neq b]\cdot\mathbf{Y}_i\cdot\hat{\beta}_{b_i} \right \| _1,
\end{equation}
where $\hat{\bm{\beta}}_{b}$ does not require the supervision of $\tilde{\bm{\beta}}_{b}$ learned from Eq.~\ref{eq:gt_beta}. 

The two proposed pretext tasks are based on learning the inherent spectral structure, but they have different focuses. S$^3$R-CR tries to figure out the similarity between $\mathcal{Y}_{\partial_b}$ and $\mathbf{Y}_b$, \emph{i.e.}, $\hat{\bm{\beta}}_b$, and focuses more on the pathological characteristics of different morphologies. S$^3$R-BR could be regarded as a novel MIM method, which focuses more on the holistic semantics of HSIs to recover $\mathbf{Y}_b$ and $\hat{\bm{\beta}}_b$ is only regarded as a latent variable.

\section{Experiments}

\subsection{Experimental Setup}\label{sec:3.1}
\subsubsection{Datasets} We verify the effectiveness of the proposed S$^3$R on two hyperspectral histopathology image datasets, in which all histopathology samples are collected from the gastroenterology department of a hospital. During the process of capturing HSI, the light transmitted from the tissue slice was collected by the microscope with the objective lens of 20$\times$. More details  are as follows: 

\paragraph{PDAC Dataset:}
It consists of 523 HSI scenes from 36 patients, which are split into 331 for training, 101 for validation, and 91 for testing. Noted that there is no overlap of patients between different splits. Among all the scenes, 255 of them belong to pancreatic ductal adenocarcinoma (PDAC), and the rest ones are normal. The wavelength is from 450 nm to 750 nm, which ends up with 40 spectral bands for each scene. The image size per band is 512 $\times$ 612. 

\paragraph{PLGC Dataset:}
Clinically, intestinal metaplasia (IM) and dysplasia (DYS) are considered as precancerous lesions of gastric cancer (PLGC) \cite{Gullo2020PLGC} and there are 1105 HSI scenes (414 IM, 362 DYS and 329 normal cases) in the dataset. All samples are randomly split into 661 for training, 221 for validation, and 223 for testing. The wavelength is from 470 nm to 670 nm, which ends up with 32 spectral bands for each scene. The image size per band is 512 $\times$ 512. 

\subsubsection{Implementation Details}

\paragraph{Pre-Training:} The pre-training process of all self-supervised algorithms is conducted on the training set, without any data augmentation. We adopt the ImageNet pre-trained model for all experiments unless otherwise specified. We set the initial learning rate to be $10^{-4}$. The maximum training epoch for all the models are 200, and early stop strategy is adopted when training loss  no longer drops. We use exponential learning rate decay with $\gamma = 0.99$. The learning rate, maximum training epochs, and learning rate decay for training are the same for other competing methods.

\paragraph{Downstream Task:} All pre-trained models can be applied in downstream tasks directly after adjusting the first convolutional layer. During fine tuning, appropriate data augmentation (\emph{e.g.}, scale rotation or gaussian noise) is added to training set. We evaluate the model on validation set after every epoch, and save the parameters when it performs best in current stage. At last, we measure the performance on the test set. Training configurations are consistent throughout the fine-tuning process to ensure fair comparisons. We use AdamW \cite{Loshchilov2017AdamW} as our optimizer and the learning rate is $10^{-4}$ with linear decline strategy. The maximum number of fine-tune epoch is 70. All experiments are implemented using Pytorch and conducted on a single NVIDIA GeForce RTX 3090. 

\begin{table}[t]
\renewcommand\arraystretch{0.8}
\setlength{\tabcolsep}{15pt}
\footnotesize
\centering
\caption{Ablation study on PLGC dataset.}
\label{tab:ablation}
\begin{tabular}{ccc|cc}
\toprule[0.15em]
Image masking & S$^3$-BR & S$^3$-CR & Backbone & Acc.\\
\midrule
× & × & × & ResNet18 & 94.17\\
\checkmark & × & × & ResNet18 & 93.72\\
\checkmark & \checkmark & × & ResNet18 & 94.62\\
\checkmark & × & \checkmark & ResNet18 & \textbf{95.07}\\
\arrayrulecolor{black!30}\midrule
× & × & × &  ResNet50 & 95.07\\
\checkmark & × & × & ResNet50 & 94.17\\
\checkmark & \checkmark & × & ResNet50 & 95.52\\
\checkmark & × & \checkmark & ResNet50 & \textbf{96.41}\\
\arrayrulecolor{black}\bottomrule[0.15em]
\end{tabular}
\end{table}

\begin{figure}[!tb]
\begin{center}
\includegraphics[width=0.9\textwidth]{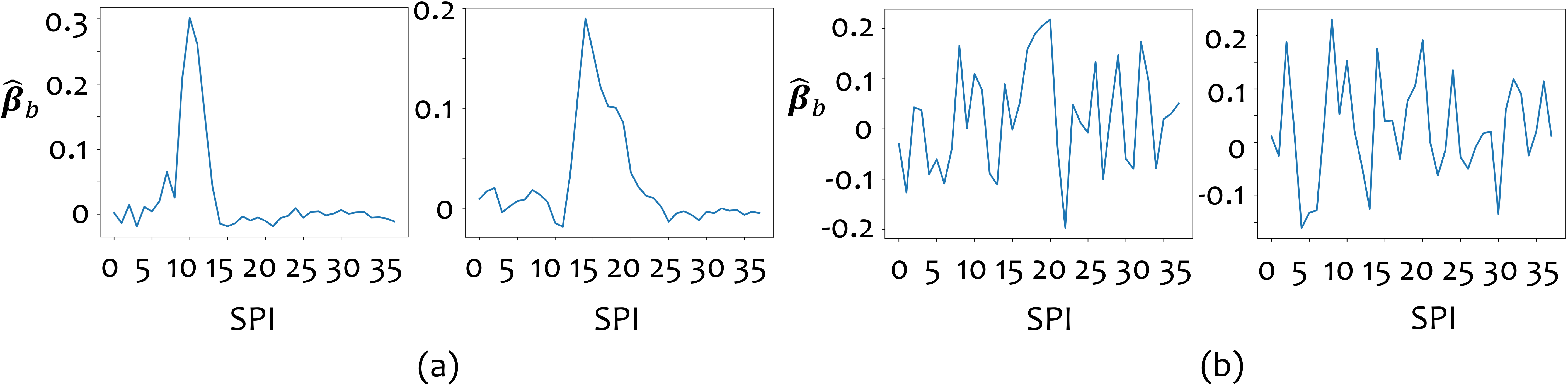}
\end{center}
\caption{Visualizations of $\hat{\bm{\beta}}_b$. The images in figure (a) come from S$^3$R-CR, and the ones in figure (b) are from S$^3$R-BR. SPI indicates spectral band index. The regression targets from left to right are 10th and 15th band in both figures.} \label{fig3}
\end{figure}

\subsection{Ablation Study\label{sec:3.2}}

The ablations are conducted on PLGC dataset with ResNet and the same pre-trained setup mentioned in sect.~\ref{sec:3.1}. As shown in Table~\ref{tab:ablation}, we first evaluate a straightforward regression method. $\mathcal{Y}_{\partial _b}$ is first fed into a vanilla ResNet backbone. Next, a decoder consisting of five transposed convolutional layers is implemented to regress the target band $\mathbf{Y}_b$. This strategy only obtains 94.17\% and 95.07\% accuracy with two backbones. While masking the input HSI, the performance gets worse. This may due to the reason that directly regressing the missing band by feeding the remaining bands to a network does not use the inherent structure of HSIs. Thus, it leads to worse performance.

\subsection{Comparison between S$^3$R and Other Methods \label{sec:3.3}}
In this section, we conduct comparisons between S$^3$R
and three competitors: 1) contrastive learning (BYOL and SimSiam), 2) self-supervised models designed for remote sensing HSIs (Conv-Deconv \cite{Mou2018Conv-Deconv}) and SSAD \cite{Yue2022SSAD}), and 3) an MIM-like method \cite{Xie2021SimMIM}, termed as Masked HyperSpectral Image Modeling (MHSIM). Vanilla siamese structure based on contrastive learning (\emph{e.g.}, BYOL and SimSiam) is designed for natural images. Thus, we use two strategies to handle the input: randomly selecting 3 bands from HSIs or using all bands. For the MIM-like method, we first randomly mask out a portion of the input, and then reconstruct the missing pixels. Noted that DINO \cite{Caron2021DINO} and other transformer-based methods \cite{He2021MOCOv3} need to be trained on large scale datasets, such as ImageNet. Our dataset contains only hundreds of HSIs. Moreover, due to the lack of computing resources, the batch size for training DINO is 25, which is far from enough. Thus, it is not suitable to compare with transformer-based self-supervised learning methods.

\begin{table}[!tb]
\renewcommand\arraystretch{0.75}
\setlength\tabcolsep{8pt}
\footnotesize
\centering
\caption{Performance comparison in Classification Accuracy ($\%$) and Standard Deviation on our two datasets. Scratch means training-from-scratch and IN1K denotes ImageNet-1K. MHSIM is an MIM-like method mentioned in Sect.~\ref{sec:3.3}.}
\label{tab1}
\begin{tabular}{lccc}
\toprule[0.15em]
Method & Dataset &  ResNet18 & ResNet50\\ 
\midrule[0.09em]
Scratch  &PDAC & 78.02 (2.28)& 80.22 (1.10)\\
IN1K pre-trained  &PDAC & 83.52 (1.68)& 85.71 (1.68)\\
BYOL \cite{Jean2020BYOL} (3 bands)  & PDAC & 82.42 (1.68)& 86.81 (2.28)\\
BYOL \cite{Jean2020BYOL} (40 bands)  & PDAC & 83.52 (1.68) & 85.71 (1.68)\\
SimSiam \cite{Chen2021SimSiam} (3 bands)  & PDAC & 86.81 (1.10)& 86.81 (1.68)\\
SimSiam \cite{Chen2021SimSiam} (40 bands)  & PDAC & 85.71 (1.67)& 87.91 (1.27)\\
SSAD \cite{Yue2022SSAD} & PDAC & 80.22 (2.29)& 80.22 (2.20)\\
Conv–Deconv \cite{Mou2018Conv-Deconv}  & PDAC & 79.12 (1.68)& 80.22 (1.68)\\
\arrayrulecolor{black!30}\midrule
MHSIM & PDAC & 81.32 (1.10)& 83.52 (0.64)\\
S$^3$R-CR & PDAC & 91.21 (2.23)& \textbf{91.21(1.68)}\\
S$^3$R-BR & PDAC & \textbf{92.31(1.10)} & 90.11 (1.27)\\
\arrayrulecolor{black}\midrule\midrule
Scratch  &PLGC & 85.20 (1.58)& 84.34 (2.87)\\
IN1K pre-trained  &PLGC & 93.27 (0.51)& 94.62 (1.70)\\
BYOL \cite{Jean2020BYOL} (3 bands)  & PLGC & 94.62 (0.90)& 95.07 (0.69)\\
BYOL \cite{Jean2020BYOL} (32 bands)  & PLGC & 94.62 (0.26) & 94.17 (0.52)\\
SimSiam \cite{Chen2021SimSiam} (3 bands)  & PLGC & 94.62 (0.63) & 95.07 (0.90)\\
SimSiam \cite{Chen2021SimSiam} (32 bands)  & PLGC & 93.72 (0.69) & 92.38 (0.51)\\
SSAD \cite{Yue2022SSAD} & PLGC & 92.38 (2.99) & 91.03 (0.68)\\
Conv–Deconv \cite{Mou2018Conv-Deconv} & PLGC & 90.13 (0.93) & 90.13 (0.45)\\
\arrayrulecolor{black!30}\midrule
MHSIM & PLGC & 93.27 (1.44)& 94.17 (0.52)\\
S$^3$R-CR & PLGC & \textbf{95.07(0.21)} & \textbf{96.41(0.51)}\\
S$^3$R-BR & PLGC & 94.62 (0.42) & 95.52 (0.26)\\
\arrayrulecolor{black}\bottomrule[0.15em]
\end{tabular}
\end{table}

As shown in Table~\ref{tab1}, our S$^3$R performs significantly better than MHSIM and other contrastive learning based methods. In particular, on PDAC dataset, S$^3$R-BR outperforms BYOL and SimSiam by 8.79\% and 5.5\% in classification accuracy with the ResNet18 backbone. On PLGC dataset, S$^3$R-CR with ResNet50 backbone achieves best results. We can observe that, the performance of MHSIM is close to ImageNet pre-training, which is much lower than ours. This may caused by information leakage from CNN architecture in MIM method \cite{Fang2022corrupted}. Restricted by computing resources, contrastive learning based methods require far more pre-training time (more than 8 minutes per epoch on PLGC) than S$^3$R (about 2 minutes per  epoch on PLGC), even with 3-band image as input. Thus, our proposed method can effectively help improve the performance in HSI classification task with lower cost.

We also visualize $\hat{\bm{\beta}}{_b}$ in Fig.~\ref{fig3} to further explore our method. We can see that, in Fig.~\ref{fig3} (a), with the coefficient regression, $\hat{\bm{\beta}}_b$ exhibits a Gaussian-like distribution, which means restoring $b$th band (peak area in the figure) will be more dependent on its nearby bands. This may force the model to focus more on the inherent structures of HSIs. As shown in Fig.~\ref{fig3} (b), $\hat{\bm{\beta}}_b$ learned as latent variable does not present a similar distribution, which illustrates that using pixel-level band regression as the target makes the encoder to extract features by the holistic semantics rather than detailed morphologies.

\section{Conclusion}
We first attempt to address the problem of self-supervised pre-training for hyperspectral histopathology image classification. We present self-supervised spectral regression (S$^3$R), by exploring the low rankness in the spectral domain of an HSI. We assume one spectral band can be approximately represented by the linear combination of the remaining bands. Our S$^3$R forces the network to understand the inherent structures of HSIs. Intensive experiments are conducted on two hyperspectral histopathology image datasets. Experimental results show that the superiority of the proposed S$^3$R lies in both performance and training efficiency, compared with state-of-the-art self-supervised methods in computer vision.

\subsubsection{Acknowledgements} This work was supported in part by the National Natural Science Foundation of China under Grant 62101191, and in part by Shanghai Natural Science Foundation under Grant 21ZR1420800.

%
%
\bibliographystyle{splncs04}
\bibliography{mybib.bib}
\end{document}